\documentclass[conference]{IEEEtran}
\IEEEoverridecommandlockouts

\usepackage{cite}
\usepackage{amsmath,amssymb,bm}
\usepackage{algorithmic}
\usepackage[ruled]{algorithm2e}
\usepackage{graphicx}
\usepackage{textcomp}
\usepackage{mathpazo}
\usepackage{txfonts}
\usepackage{upgreek}
\usepackage{bbding}
\usepackage{adjustbox}
\usepackage[colorlinks,citecolor=black,bookmarks=false]{hyperref}

\usepackage{url}
\usepackage{array}
\usepackage{multirow}
\usepackage{adjustbox}
\usepackage{fancyvrb}
\usepackage{threeparttable}
\usepackage{color}

\def\BibTeX{{\rm B\kern-.05em{\sc i\kern-.025em b}\kern-.08em
    T\kern-.1667em\lower.7ex\hbox{E}\kern-.125emX}}
\providecommand{\tabularnewline}{\\}

\begin{document}
\title{Parameter Transfer Extreme Learning Machine based on Projective Model\\
\thanks{This work was supported by the National Science and Technology Major Project (No. 2013ZX03005013), and the opening foundation of the State Key Laboratory (NO. 2014KF06).}
}

\author{\IEEEauthorblockN{ Chao Chen}
\IEEEauthorblockA{\textit{Institute of Information Science} \\  \textit{and Electronic Engineering}\\
\textit{Zhejiang University}\\
Hangzhou, China \\
chench@zju.edu.cn}
\and
\IEEEauthorblockN{Boyuan Jiang}
\IEEEauthorblockA{\textit{Institute of Information Science} \\  \textit{and Electronic Engineering}\\
\textit{Zhejiang University}\\
Hangzhou, China \\
byjiang@zju.edu.cn}
\and
\IEEEauthorblockN{Xinyu Jin\Envelope}
\IEEEauthorblockA{\textit{Institute of Information Science} \\  \textit{and Electronic Engineering}\\
\textit{Zhejiang University}\\
Hangzhou, China \\
jinxy@zju.edu.cn}


}
\maketitle

\begin{abstract}
Recent years, transfer learning has attracted much attention in the community of machine learning. In this paper, we mainly focus on the tasks of parameter transfer under the framework of extreme learning machine (ELM). Unlike the existing parameter transfer approaches, which incorporate the source model information into the target by regularizing the difference between the source and target domain parameters, an intuitively appealing projective-model is proposed to bridge the source and target model parameters. Specifically, we formulate the parameter transfer in the ELM networks by the means of parameter projection, and train the model by optimizing the projection matrix and classifier parameters jointly. Further more,  the $\ell_{2,1}\text{-norm}$ structured sparsity penalty is imposed on the source domain parameters, which encourages the joint feature selection and parameter transfer. To evaluate the effectiveness of the proposed method, comprehensive experiments on several commonly used domain adaptation datasets are presented. The results show that the proposed method significantly outperforms the non-transfer ELM networks and other classical transfer learning methods.
\end{abstract}


\section{Introduction}
\begin{figure}[tbp]
\begin{center}
\maxsizebox{\columnwidth}{!}{\includegraphics{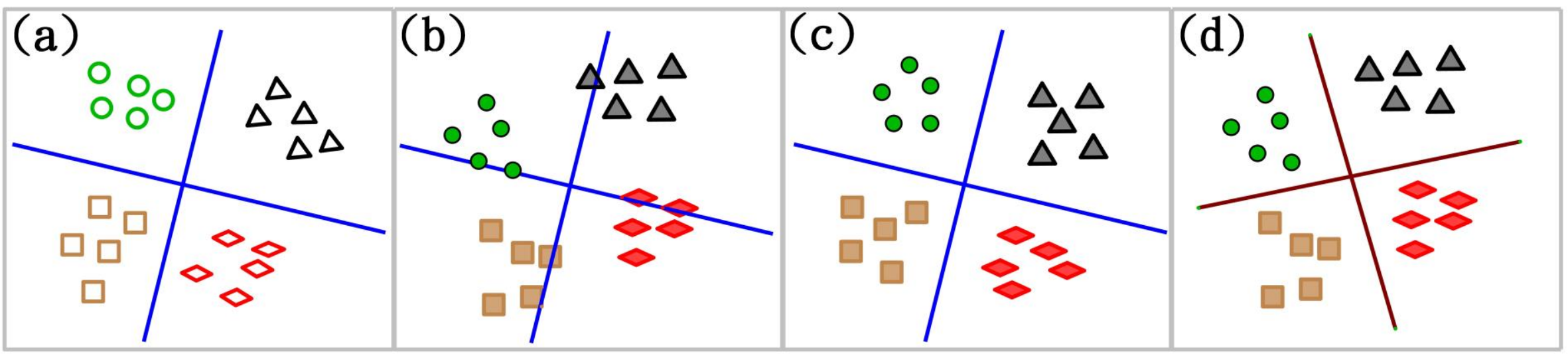}}
\end{center}
\caption{Illustration of the transform-based approaches $\mathrm{v.s.}$ parameter transfer approaches in a four-class classification problem. (a) The hyperplane learned in the source domain. (b) Use the hyperplane learned in the source domain on the target domain directly. Due to the domain-shift, several samples are misclassified. (c) Correct the domain-shift with the transform-based approaches first, then the learned hyperplane is used for classification. (d)  Adapting the learned hyperplane to the target domain with a small number of target instances.}
\label{Fig0}
\end{figure}

\indent In traditional machine learning and pattern classification methods, there is a strong assumption that all the data are drawn from the same distribution. However, this assumption may not always hold in many real world scenarios. For example, in cases where the training samples are difficult or expensive to obtain, or when the distribution of the samples changes over time, we have to borrow knowledge from another different but highly related domain. Therefore, how to transfer knowledge from another different but related domain has become more and more important. During the past two decades, transfer learning has emerged as a new framework to solve this problem, and has received more and more attention in the machine learning and data mining community. As has been discussed in\cite{pan2010survey}, feature matching based methods are the most widely used transfer learning approaches, which aim to learn a shared feature representation to minimize the distribution discrepancy between the source and target domain \cite{long2014transfer,long2014adaptation,zhang2016robust,liu2017common,hoffman2014asymmetric,saenko2010adapting,gong2012geodesic,hoffman2013efficient,kulis2011you,sun2016return,chen2018joint}. Among them, to learn a cross-domain transformations that maps the target features into the source\cite{saenko2010adapting,kulis2011you,gong2012geodesic,hoffman2013efficient,hoffman2014asymmetric,sun2016return} is of great importance. Apart from this, the parameter transfer approach is another highly concerned line of works. It assumes that the transferred knowledge has been encoded into the hyper-parameters of the classification model\cite{yang2007adapting,aytar2011tabula,tommasi2014learning,li2016extreme}. Therefore, the source model and target model should share some parameters or prior distribution of the model parameters. Based on this assumption, parameter transfer approaches could adapt the learned source hyperplane to the target domain with a small number of target samples. As illustrated in Figure \ref{Fig0}, we show the comparison of the transform-based methods and parameter transfer methods. The transform-based methods map the features to adapt to the learned hyperplane, while the parameter transfer approaches adjust the learned hyperplane to adapt to the shifted features. Despite that a large number of transform-based methods and parameter transfer methods have been proposed to address the knowledge transfer problem, few works have tried to combine these two methods together. In this paper, we attempt to comply with parameter transfer based on the projective-model, especially under the framework of the extreme learning machine.

As a special feed-forward neural networks, the Extreme learning machine (ELM) first proposed by Huang\cite{huang2006extreme}, which determines its input weights randomly, has become a very popular classifier due to its fast learning speed, satisfactory performance and little human intervention\cite{huang2012extreme}. Therefore, since its first appearance, various extensions have been proposed to make the original ELM model more efficient and suitable for specific applications. Chen et al. optimized the input weights of ELM by generalized hebbian learning and intrinsic plasticity learning \cite{chen2018optimizing}. Based on the manifold regularization framework, Huang et al. extended ELM to semi-supervised and unsupervised learning in \cite{huang2014semi}. To handle the imbalanced data problem, Zong et al. extended the traditional ELM to the weighted ELM (WELM) in \cite{zong2013weighted}. Considering the computational cost and spatial requirements, the online version of ELM has also been proposed and studied\cite{liang2006fast}. Besides, several multi-layer ELM frameworks \cite{zhou2015stacked,huang2015local} have also been put forward  recently to learn deep representation of the original data.

In this paper, we mainly focus on the parameter transfer approach based on the ELM algorithm. We would like to learn a high-quality ELM classifier using a small number of labeled target domain samples and a large number of source domain samples. To achieve this goal, we assume that the source domain classification hyperplane and the target domain classification hyperplane could be bridged by a projection matrix, i.e. the target domain parameters can be represented as a projection matrix multiplying with the source domain parameters. In this way, the parameter transfer ELM model can be learned by jointly optimizing the ELM model parameters and the projection matrix. Further more, the $\ell_{2,1}\text{-norm}$ of the source domain parameters is incorporated into the objective function, which leads to selecting useful features in the source domain during model training. For ease of notation, the proposed parameter transfer ELM is referred to as PTELM.

The contributions of this paper are four-fold: Firstly, we are among the first to exploit projective-based model for parameter transfer, especially under the framework of the ELM. Secondly, unlike most of existing works which learn the transformations by minimizing the distribution discrepancy or maximizing some kind of similarity metric between the source and target feature space, the proposed PTELM jointly learns the projection matrix and the model parameter by minimizing the classification error directly. Thirdly, the $\ell_{2,1}\text{-norm}$ is imposed on the source domain hyperplane. In this respect, the learned source model tends to select informative features for knowledge transfer. Lastly, we demonstrate that the proposed parameter transfer ELM can also be regarded as a special transform-based domain adaptation method.

\section{Related Works}
Recently, some researchers have focused their attention on the domain adaptation ELM. Zhang et al. proposed a domain adaptation ELM to address the sensor drift problem in the E-nose system\cite{zhang2015domain}. In \cite{liu2017common}, a unified subspace transfer framework based on ELM was proposed, which learns a subspace that jointly minimizes the mean distribution discrepancy (MMD) and maximum margin criterion (MMC). Uzair et al. \cite{uzair2017blind} proposed a blind domain adaptation ELM with extreme learning machine auto-encoder (ELM-AE), which does not need target domain samples for training. In\cite{zhang2016robust}, Zhang et al. proposed a ELM-based domain adaptation (EDA) for visual knowledge transfer and extended the EDA to multi-view learning. In EDA, the manifold regularization was incorporated into the objective function, and the author minimizes the $\ell_{2,1}\text{-norm}$ of the hyperplane and prediction error simultaneously. Besides, a parameter transfer approach based transfer learning ELM (TLELM) has also been proposed in \cite{li2016extreme}, which regularizes the difference between the source and target parameters. In addition, Salaken et al.\cite{salaken2017extreme} summarized all the available literatures in the filed of ELM based transfer learning methods.

Among the parameter transfer approaches, the majority of related works incorporated the source model information into the target by regularizing the difference of the parameters between the source and the target domain\cite{yang2007adapting,aytar2011tabula,tommasi2014learning,li2016extreme}. The representative method is the adaptive SVM (A-SVM)\cite{yang2007adapting}, which learns from the source domain parameters by directly regularizing the distance between the learned model and the target model.  After that, Aytar et al. \cite{aytar2011tabula} proposed two new parameter transfer SVM, which extends and relaxes the A-SVM. Li et al.\cite{li2016extreme} proposed transfer learning ELM by introducing the same regularizer as A-SVM in the ELM.

\section{Preliminaries}
\subsection{A Brief Review of ELM}
\indent Considering a supervised learning problem where the training set with $N$ samples and the corresponding targets are given as $\{\mathbf{X,Y}\}=\{(\bm{x}_i,\bm{y}_i)|\bm{x}_i\in \mathbb{R}^n,\bm{y}_i\in \mathbb{R}^m,i=1,2,\cdots,N\}$. Here $\bm{x}_i\in \mathbb{R}^n$ is the n-dimensional input data and $\bm{y}_i\in \mathbb{R}^m$ is its associated one-hot labels. The ELM networks learns a decision rule with the following two stages. In the first stage, it randomly generates the input weights $\bm{w}$ and bias $\bm{b}$, and maps the original data from the input space into the $L$-dimensional feature space $\bm{h}(\bm{x}_i)\in \mathbb{R}^L$, where $L$ is the number of hidden nodes, $\bm{h}(\bm{x}_i)=\bm{g}(\bm{w}^{\top}\bm{x}_i+b)$, and $\bm{g(\cdot)}$ is the activation function. In this respect, the only free parameter of the ELM is the output weights $\bm{\betaup}\in \mathbb{R}^{L\times m}$. In the second stage, the ELM solves the output weights by minimizing the square loss summation of prediction errors and the norm of the output weights simultaneously, leads to
\begin{equation}
\begin{cases}
\min\limits_{\bm{\betaup}} \mathcal{L}(\bm{\betaup})=\dfrac{1}{2}\Vert\bm{\betaup}\Vert_F^2+\tfrac{\lambda}{2}\sum\limits_{i=1}^N\Vert\bm{\xi}_i\Vert_2^2 \\
\mathrm{s.t.} \quad \bm{h}(\bm{x}_i)\bm{\betaup}=\bm{y}_i-\bm{\xi}_i, i=1,2,\cdots,N
\end{cases}
\end{equation}
where $\bm\xi_i$ is the prediction error with respect to the $i$-th training sample, the first term of the objective function is the regularization term to prevent the network from overfitting. By substituting the constrain into the objective function, the problem (1) can be simplified to such an unconstrain optimization problem:
\begin{equation}
\min\limits_{\bm{\betaup}} \mathcal{L}(\bm{\betaup})=\dfrac{1}{2}\Vert\bm{\betaup}\Vert_F^2+\dfrac{\lambda}{2}\Vert\mathbf{H}\bm{\betaup}-\mathbf{Y}\Vert_F^2
\end{equation}
where $\mathbf{H}=[\bm{h}(\bm{x}_1);\bm{h}(\bm{x}_2);\cdots;\bm{h}(\bm{x}_N)]\in \mathbb{R}^{N\times L}$. The optimal solution of $\bm{\betaup}$ can then be analytically determined by setting the derivatives of $\mathcal{L}(\bm{\betaup})$ with respect to $\bm{\betaup}$ to zero, i.e.
\begin{equation}
\frac{\partial \mathcal{L}(\bm{\betaup})}{\partial \bm{\betaup}}=\bm{\betaup}+\lambda\mathbf{H}^{\top}(\mathbf{H}\bm{\betaup}-\mathbf{Y})=0
\end{equation}
Then, the output weights $\bm{\betaup}$ can be effectively solved by
\begin{equation}
\bm{\betaup}=(\mathbf{H}^{\top}\mathbf{H}+\dfrac{\mathbf{I}}{\lambda})^{-1}\mathbf{H}^{\top}\mathbf{Y}
\end{equation}
where $\mathbf{I}$ is the identity matrix and $\lambda$ is the regularization coefficient. With the closed-form solution, the ELM model is remarkably efficient and tends to reach a global optimum.

\subsection{Notations and Definitions}
 We summarize the frequently used notations and definitions as below.
 \\\indent \textbf{Notations:} For a matrix $\mathbf{A}\in\mathbb{R}^{m\times n}$, let the $i$-th row of $\mathbf{A}$ denoted by $\mathbf{a}^i$. The Frobenius norm of the matrix $\mathbf{A}$ is defined as
 \begin{equation}
\Vert\mathbf{A}\Vert_F=\sqrt{\sum\limits_{i=1}^m\sum\limits_{j=1}^n\mathrm{a}_{ij}^2}=\sqrt{\sum\limits_{i=1}^{m}\Vert\mathbf{a}^i\Vert_2^2}
 \end{equation}
 The $\ell_{2,1}\text{-norm}$ of a matrix, introduced in \cite{ding2006r} firstly as rotation invariant $\ell_1\text{-norm}$ which ensures the row sparsity of a matrix, was widely used for feature selection and structured sparsity regularizer\cite{gu2011joint,nie2010efficient,long2014transfer,zhang2016robust}. It is defined as
 \begin{equation}
\Vert\mathbf{A}\Vert_{2,1}=\sum\limits_{i=1}^m\sqrt{\sum\limits_{j=1}^n\mathrm{a}_{ij}^2}=\sum\limits_{i=1}^{m}\Vert\mathbf{a}^i\Vert_2
 \end{equation}
 \\\textbf{Definition 1. Domain.} A domain $\mathcal{D}$ is composed of a feature space $\mathcal{X}$ and a marginal distribution $\mathcal{P(X)}$. $\mathcal{D}_s=\{\mathcal{X}_s,\mathcal{P}(\mathcal{X}_s)\}$ and $\mathcal{D}_t=\{\mathcal{X}_t,\mathcal{P}(\mathcal{X}_t)\}$ represent the source and target domain respectively, which are sampled from different but related distributions. Generally, $\mathcal{X}_s\neq \mathcal{X}_t$ and $\mathcal{P}(\mathcal{X}_s)\neq \mathcal{P}(\mathcal{X}_t)$.
\\\textbf{Definition 2. Transfer Learning.} For the given source domain $\mathcal{D}_s=\{(\bm{x}_1,\bm{y}_1),\cdots,(\bm{x}_m,\bm{y}_m)\}$, and target domain
$\mathcal{D}_t=\{(\bm{x}_1,\bm{y}_1),\cdots,(\bm{x}_n,\bm{y}_n)\}$. Generally, $m\gg n$. Data in the target domain are insufficient to learn a high-quality classification model. Transfer learning aims to learn a satisfied classifier with the incorporation of the source domain information.
\\\textbf{Definition 3. Parameter Transfer.} Define $\Theta_s$ and $\Theta_t$ are the model parameters learned from the two domains.
\begin{equation}
\Theta_{s(t)}^*=\mathop{\arg\min}\limits_{\Theta_{s(t)}}\sum\limits_{\bm{x}_i\in\mathcal{X}_{s(t)}}\mathcal{L}(\bm{y}_i,\mathcal{T}(\bm{x}_i;\Theta_{s(t)}))+\gamma\mathcal{R}(\Theta_{s(t)})
\end{equation}
 where the first item is the loss function and the second item is the parameter regularization. $\bm{f}_s(\bm{x})=\mathcal{T}(\bm{x};\Theta_s)$ is the classification model learned from the source domain $\mathcal{D}_s$, and $\bm{f}_t(\bm{x})=\mathcal{T}(\bm{x};\Theta_t)$ is the classification model learned from the target domain $\mathcal{D}_t$. Based on the assumption that $\Theta_s$ and $\Theta_t$ should share some parameters or prior distribution, parameter transfer learning aims to transfer knowledge from the $\Theta_s$ to improve the target domain classification model.
 \begin{equation}
 \small
\Theta_t^*=\mathop{\arg\min}\limits_{\Theta_t}\sum\limits_{\bm{x}_i\in\mathcal{X}_t}\mathcal{L}(\bm{y}_i,\mathcal{T}(\bm{x}_i;\Theta_t))+\gamma_1\mathcal{R}(\Theta_t)+\gamma_2\mathcal{R}(\Theta_s,\Theta_t)
\small
\end{equation}
The last item $\mathcal{R}(\Theta_s,\Theta_t)$ tries to incorporate the information of the $\Theta_s$ into the $\Theta_t$.
\\\indent Most parameter transfer approaches seek to leverage the target model by the discrepancy between $\Theta_s$ and $\Theta_t$, i.e. $\mathcal{R}(\Theta_s,\Theta_t)=\Vert\Theta_s-\Theta_t\Vert^2$.
This penalty directly regularizing the distance between $\Theta_t$ and $\Theta_s$ is too strict sometimes. When $\gamma_2$ is large enough, it leads to $\Theta_t=\Theta_s$. In order to relax this  constraint, in this paper, we propose the projective-model based parameter transfer approach to bridge the source and target domain parameters.
\\\textbf{Definition 4. Projective-Model based Parameter Transfer.} Define a projection matrix $\mathcal{M}$, the projective-model based parameter transfer assumes that the source domain parameters and target domain parameters could be bridged by $\Theta_t=\mathcal{M}\Theta_s$.

\section{Proposed Method}
\indent In this section, we present the proposed projective-model based parameter transfer ELM and its learning algorithm.
\subsection{Problem Formulation}
Suppose we have a source domain with $m$ labeled samples $\mathcal{D}_s=\{(\bm{x}_s^1,\bm{y}_s^1),\cdots,(\bm{x}_s^m,\bm{y}_s^m)\}$, and a target domain with $n$ labeled samples $\mathcal{D}_t=\{(\bm{x}_t^1,\bm{y}_t^1),\cdots,(\bm{x}_t^n,\bm{y}_t^n)\}$. Generally, in domain adaptation algorithm, $n$ is a very small number and $m\gg n$. Denoting $\bm{\upbeta}_s$ and $\bm{\upbeta}_t$ as the source and target ELM model parameters need to be optimized. As discussed above, we tend to bridge the source domain parameters and the target parameters by a projection matrix $\mathbf{M}$, i.e. $\bm{\upbeta}_t=\mathbf{M}\bm{\upbeta}_s$. Our goal is to learn the ELM classification hyperplane and the projection matrix jointly. In this respect, the objective function can be formulated as
\begin{equation}
\min\limits_{\bm{\upbeta}_s,\mathbf{M}}\dfrac{1}{2}\sum_{j=1}^n\Vert\bm{\xi}_t^j\Vert^2+\dfrac{\lambda_1}{2}\sum_{i=1}^m\Vert\bm{\xi}_s^i\Vert^2+\dfrac{\lambda_2}{2}\Vert\bm{\upbeta}_s\Vert_{2,1}+\dfrac{\lambda_3}{2}\Vert\bm{\upbeta}_t\Vert_F^2
\end{equation}
\begin{equation}
\mathrm{s.t.}\
\begin{cases}
\bm{h}_s(\bm{x}_s^i)\bm{\betaup}_s=\bm{y}_s^i-\bm{\xi}_s^i, i=1,2,\cdots,m \qquad\qquad\qquad\\
\bm{h}_t(\bm{x}_t^j)\bm{\betaup}_t=\bm{y}_t^j-\bm{\xi}_t^j, j=1,2,\cdots,n \qquad\qquad \\
\bm{\upbeta}_t=\mathbf{M}\bm{\upbeta}_s  \qquad\qquad\qquad\qquad\qquad\quad
\end{cases}
\end{equation}
where $\bm{h}_s(\bm{x}_s^i)$, $\bm{y}_s^i$ and $\bm{\xi}_s^i$ denote the outputs of the hidden layer, the one-hot label vector and the prediction error with respect to the $i$-th samples from the source domain. Similarly, $\bm{h}_t(\bm{x}_t^j)$, $\bm{y}_t^j$ and $\bm{\xi}_t^j$ denote the output of the hidden layer, the one-hot label vector and the prediction error with respect to the $j$-th samples from the target domain. As can be seen, there are four terms altogether in the objective function, which are intuitive to understand. The first two terms tend to simultaneously minimize the training error in the source and target domain, and the last two terms are used for preventing the source and target ELM model from overfitting. $\lambda_1$, $\lambda_2$, $\lambda_3$ are trade-off parameters used to balance the contributions of the four terms to the objective function. The merits that distinguish our proposal from other related works are two-fold: On the one hand, different from the traditional parameter transfer approach\cite{li2016extreme}, we bridge the source domain and the target domain parameters by a projection matrix. On the other hand, the $\ell_{2,1}\text{-norm}$ instead of the Frobenius norm is imposed on the source domain hyperplane as a regularizer. With this penalty,  the row sparsity $\bm{\upbeta}_s$ will be obtained. Benefiting from this property, our model tends to select the informative features in the source domain for knowledge transfer.
\\\indent By substituting the constrains into the objective function, the optimization function (9) can be easily reformulated as an equivalent unconstrained optimization problem.
\begin{equation}
\begin{split}
\min\limits_{\bm{\upbeta}_s,\mathbf{M}}\mathcal{L}(\bm{\upbeta}_s,\mathbf{M})=&\dfrac{1}{2}\Vert \mathbf{H}_t\mathbf{M}\bm{\upbeta}_s -\mathbf{Y}_t\Vert_F^2+\dfrac{\lambda_1}{2}\Vert \mathbf{H}_s\bm{\upbeta}_s-\mathbf{Y}_s\Vert_F^2\\
+&\dfrac{\lambda_2}{2}\Vert\bm{\upbeta}_s\Vert_{2,1}+\dfrac{\lambda_3}{2}\Vert\mathbf{M}\bm{\upbeta}_s \Vert_F^2 \\
\end{split}
\end{equation}

Where $\mathbf{H}_t\in \mathbb{R}^{n\times L}$ and $\mathbf{H}_s\in \mathbb{R}^{m\times L}$ denote the hidden layer outputs of the target and the source ELM model, $\bm{\upbeta}_t\in \mathbb{R}^{L\times c}$ and $\bm{\upbeta}_s\in \mathbb{R}^{L\times c}$ denote the output weights of the target and source ELM model. $\mathbf{Y}_t\in \mathbb{R}^{n\times c}$ and $\mathbf{Y}_s\in \mathbb{R}^{m\times c}$ denote the label matrix of the target and source domain samples. Here, $L$ denotes the number of hidden nodes in the ELM model, and $c$ is the number of classes of the source and target domains.

\subsection{Learning Algorithm}
As can be seen in problem (11), our goal is to jointly learn the output weights of the source ELM model $\bm{\upbeta}_s$ and the projection matrix $\mathbf{M}$. Then, the target ELM model parameters can be easily obtained by $\bm{\upbeta}_t=\mathbf{M}\bm{\upbeta}_s$. However, with two free parameters to be solved, this optimization problem can not be directly solved like the problem (2). Therefore, we adopt the coordinate descent method to alternatively optimize the two free parameters.
\\\indent \textbf{(1) Fix $\mathbf{M}$ and optimize on $\bm\upbeta_s$:} In the first step, we fix the projection matrix as $\mathbf{M}^{(0)}=\mathbf{I}$, then, the sub-problem $\bm\upbeta_s^*=\arg\min_{\bm\upbeta_s}\mathcal{L}(\bm\upbeta_s,\mathbf{M})$ can be solved by setting the derivative of objective function $\mathrm{w.r.t}$ $\bm\upbeta_s$ to zero. Then we have
\begin{equation}
\begin{split}
\dfrac{\partial\mathcal{L}(\bm\upbeta_s,\mathbf{M})}{\partial\bm\upbeta_s}=&\mathbf{M}^{\top}\mathbf{H}_t^{\top}(\mathbf{H}_t\mathbf{M}\bm\upbeta_s-\mathbf{Y}_t)+\lambda_1\mathbf{H}_s^{\top}(\mathbf{H}_s\bm\upbeta_s-\mathbf{Y}_s)\\
&+\lambda_2\mathbf{D}\bm\upbeta_s+\lambda_3\mathbf{M}^{\top}\mathbf{M}\bm\upbeta_s=0
\end{split}
\end{equation}
Note that $\Vert\bm\upbeta_s\Vert_{2,1}$ is a non-smooth function at zero, therefore, we compute its sub-gradient instead. $\tfrac{\partial\Vert\bm\upbeta_s\Vert_{2,1}}{\partial\bm\upbeta_s}=2\mathbf{D}\bm\upbeta_s$, where $\mathbf{D}$ is a diagonal sub-gradient matrix with the $i$-th element as
\begin{equation}
\mathbf{D}_{ii}=\dfrac{1}{2\Vert\bm\upbeta_s^i\Vert_2+\epsilon}
\end{equation}
Here, $\bm\upbeta_s^i$ denotes the $i$-th row of $\bm\upbeta_s$, $\epsilon$ set as a very small constant to prevent the dividend to be zero. With the fixed matrix $\mathbf{D}$, $\bm\upbeta_s$ could be solved according to Eq. (12), as
\begin{equation}
\begin{split}
\bm\upbeta_s=&(\lambda_1\mathbf{H}_s^{\top}\mathbf{H}_s+\mathbf{M}^{\top}\mathbf{H}_t^{\top}\mathbf{H}_t\mathbf{M}+\lambda_2\mathbf{D}+\lambda_3\mathbf{M}^{\top}\mathbf{M})^{-1} \\&\times(\lambda_1\mathbf{H}_s^{\top}\mathbf{Y}_s+\mathbf{M}^{\top}\mathbf{H}_t^{\top}\mathbf{Y}_t)
\end{split}
\end{equation}
Note that the sub-gradient matrix $\mathbf{D}$ is dependent on the unsolved parameters $\bm\upbeta_s$. Thus, we employ an alternate optimization strategy to solve $\bm\upbeta_s$ according to Eq. (13) and Eq. (14). In each iteration, only one parameter is updated with the other one fixed. The algorithm is summarized in algorithm 1. It is worth noting that the iterative procedure will be terminated once the number of iterations reaches $T_{max}$ or the $\bm\upbeta_s$ tends to convergence. The convergence of this algorithm can be easily proved similar to \cite{nie2010efficient}.
\begin{algorithm}
\caption{An efficient iterative algorithm to solve $\bm\upbeta_s$ }
\KwIn{$\mathbf{H}_s,\mathbf{H}_t,\mathbf{M},\mathbf{Y}_s,\mathbf{Y}_t$}
\KwOut{$\bm\upbeta_s$}
Set t=0. Initialize $\mathbf{D}^{0}$ as an identity matrix\;
\Repeat {Converges}
{
Update $\bm\upbeta_s^{t+1}$ according to Eq. (14)\;
Update $\mathbf{D}^{t+1}$ according to Eq. (13)\;
t=t+1
}
\end{algorithm}

\indent \textbf{(2) Fix $\bm\upbeta_s$ and optimize on $\mathbf{M}$:} With the fixed $\bm\upbeta_s$, the sub-problem $\mathbf{M}^*=\arg\min_{\mathbf{M}}\mathcal{L}(\bm\upbeta_s,\mathbf{M})$ can be easily solved by taking the derivative of Eq. (11) with respect to $\mathbf{M}$ to zero. We get
\begin{equation}
\dfrac{\partial\mathcal{L}(\bm\upbeta_s,\mathbf{M})}{\partial\mathbf{M}}=\mathbf{H}_t^{\top}(\mathbf{H}_t\mathbf{M}\mathbf{\bm\upbeta_s-\mathbf{Y}_t})\bm\upbeta_s^{\top}+\lambda_3\mathbf{M}\bm\upbeta_s\bm\upbeta_s^{\top}=0
\end{equation}
which leads to
\begin{equation}
\mathbf{M}=(\mathbf{H}_t^{\top}\mathbf{H}_t+\lambda_3\mathbf{I})^{-1}\mathbf{H}_t^{\top}\mathbf{Y}_t\bm\upbeta_s^{\top}(\bm\upbeta_s\bm\upbeta_s^{\top})^{-1}
\end{equation}
\indent The overall learning algorithm is summarized in Algorithm 2. With the randomly initialized input parameters, the hidden layer outputs of the source and target ELM model  , which are represented as $\mathbf{H}_s$ and $\mathbf{H}_t$, could be calculated beforehand. In each iteration, we update $\bm\upbeta_s$ with current $\mathbf{M}$, then, update $\mathbf{M}$ with the current calculated $\bm\upbeta_s$. Owning to the closed-form solutions in each iteration, the learning algorithm will converge after several iterations.
\begin{algorithm}
\caption{Learning Algorithm of the PTELM Method}
\KwIn{$\mathcal{D}_s=(\mathbf{X}_s,\mathbf{Y}_s),\mathcal{D}_t=(\mathbf{X}_t,\mathbf{Y}_t)$}
\KwOut{$\bm\upbeta_s$,$\mathbf{M}$}
Calculate $\mathbf{H}_s$ and $\mathbf{H}_t$ with random initialized input parameters\;
Set t=0. Initialize $\mathbf{M}^{0}$ as an identity matrix\;
\Repeat {Converges}
{
Update $\bm\upbeta_s^{t+1}$ according to Algorithm 1\;
Update $\mathbf{M}^{t+1}$ according to Eq. (16)\;
t=t+1
}
\end{algorithm}

\subsection{Relationship to Transform-based Methods}
Most existing domain adaptation methods apply knowledge transfer by learning a cross-domain transformations $\mathbf{A}$\cite{hoffman2013efficient,kulis2011you,hubert2016learning,sun2016return}, which maps the source domain data into the target by applying $\widetilde{\mathbf{X}}_s=\mathbf{A}\mathbf{X}_s$. Instead, our proposed PTELM aims to transform the source domain hyperplane into the target by $\bm{\upbeta}_t=\mathbf{M}\bm{\upbeta}_s$. In fact, the proposed PTELM can also be regarded as the transform-based method. As can be seen in Eq. (11), we implicitly define $\widetilde{\mathbf{H}}_t=\mathbf{H}_t\mathbf{M}$, and the $\lambda_3$ is set to be zero. Such that the objective function can be reformulated as
\begin{equation}
\begin{cases}
\min\limits_{\bm{\upbeta},\mathbf{M}}\dfrac{1}{2}\Vert\widetilde{\mathbf{H}}_t\bm{\upbeta}-\mathbf{Y}_t\Vert_F^2+\dfrac{\lambda_1}{2}\Vert \mathbf{H}_s\bm{\upbeta}-\mathbf{Y}_s\Vert_F^2+\dfrac{\lambda_2}{2}\Vert\bm{\upbeta}\Vert_{2,1} \\
\mathrm{s.t.}\quad \widetilde{\mathbf{H}}_t=\mathbf{H}_t\mathbf{M}
\end{cases}
\end{equation}
Similar to the cross-domain transformation approaches, the above rewritten objective function aims to jointly learn a transformation matrix that transforms the target feature into the source, and the classification hyperplane. The differences between our proposal and the other transform-based methods are three-fold. On the one hand, our proposed PTELM transforms the target into the source by column transformation, while the majority of transform-based methods align the source and target by applying row transformation on the source data. On the other hand, the PTELM learns the transformation directly based on the prediction error, while other related works take the distribution discrepancy or similarity metric as guidelines. Lastly, the PTELM learns the transformation and a ELM classifier simultaneously, while many of transform-based methods simply learn the transformation, and then utilize other classifiers (e.g. KNN) for classification.

\section{Experiments}
In this section, we evaluate our proposed PTELM method on several challenging real-world datasets. The source code of the PTELM is released online\footnote{\url{https://github.com/BoyuanJiang/PTELM}}.

\subsection{Datasets and Setup}
Two types of domain adaptation problems are considered: object recognition and text categorization. A summary of the properties of each domain considered in our experiments is provided in Table \ref{tab1}.

\begin{figure}[htbp]
\begin{center}
\maxsizebox{\columnwidth}{!}{\includegraphics{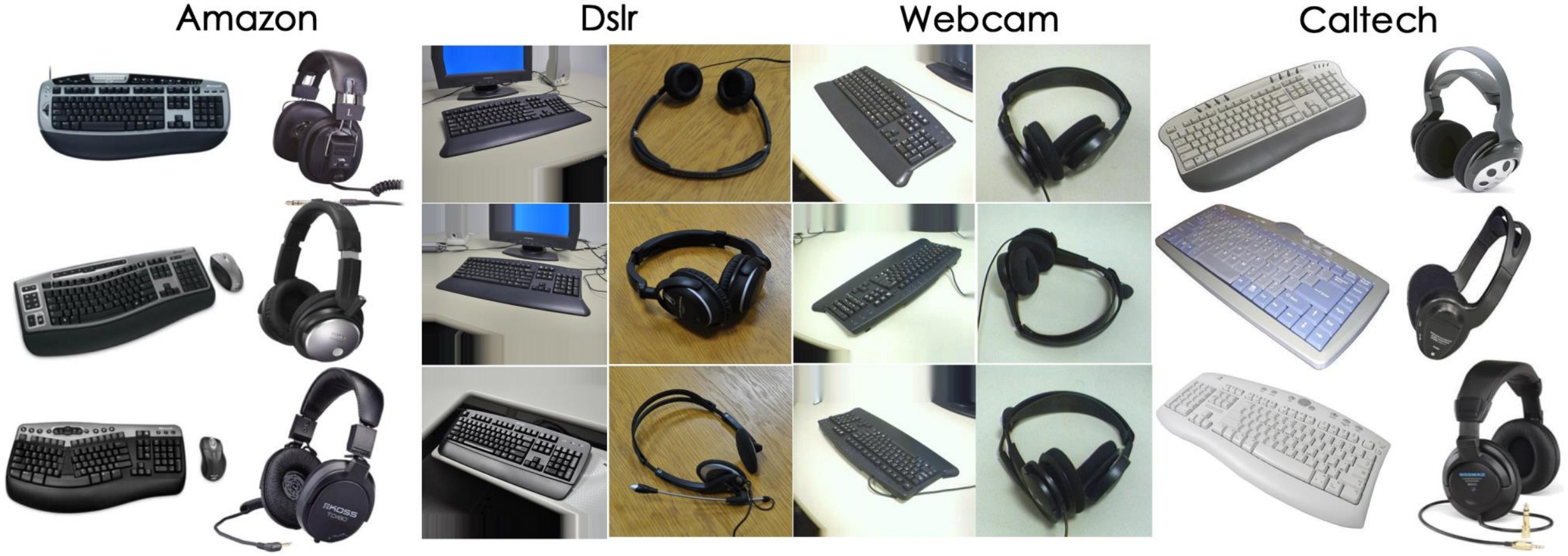}}
\end{center}
\caption{Example images of \textit{Office-Caltech} dataset. \textit{Amazon}, \textit{Dslr} and \textit{Webcam} are from \textit{Office} dataset while \textit{Caltech} is from \textit{Caltech-256} dataset. It is obvious that domain shifts are large among different domains. (Best viewed in color.)}
\label{dataset}
\end{figure}

\textbf{Caltech-Office dataset.} This dataset \cite{gong2012geodesic} consists of \emph{Office} \cite{saenko2010adapting} and \emph{Caltech-256} \cite{griffin2007caltech} datasets. It contains images from four different domains: \emph{Amazon} (product images download form amazon.com), \emph{Webcam} (low-resolution images taken by a webcam), \emph{Dslr} (high-resolution images taken by a digital SLR camera) and \emph{Caltech}. 10 common categories are extracted from all four domains with each category consisting of 8 to 151 samples, and 2533 images in total. Several factors (such as image resolution, lighting condition, noise, background and viewpoint) cause the shift of each domain. Figure \ref{dataset} highlights the differences among these domains with example images from categories of keyboards and headphones. We consider the SURF-BoW image features (\emph{SURF} in short) provided by \cite{gong2012geodesic}, which encode the images with 800-bin histograms with the codebook trained from a subset of Amazon images using SURF descriptors \cite{bay2006surf}. These histograms are then normalized to be zero means and unit variance in each dimension.

\textbf{Multilingual Reuters Collection dataset.} This dataset\footnote{\url{http://ama.liglab.fr/~amini/DataSets/Classification/Multiview/ReutersMutliLingualMultiView.htm}} \cite{AUG09,USLJ07}, which is collected by sampling from the Reuters RCV1 and RCV2 collections, contains feature characteristics of 111,740 documents originally written in five different languages and their translations (i.e., English, French, German, Italian, and Spanish), over a common set of 6 categories (i.e., C15, CCAT, E21, ECAT, GCAT, and M11). Documents belonging to more than one of the 6 categories are assigned the label of their smallest category. Therefore, there are 12-30K documents per language, and 11-34K documents per category. All documents are represented as a bag of words and then the TF-IDF features are extracted.

\textbf{Baselines} We compare the results with the following baselines and competing methods that are well adapted for domain shift scenarios:
\begin{itemize}
  \item \textbf{SVM$_s$:} Support vector machine trained on source.
  \item \textbf{SVM$_t$:} Support vector machine trained on target.
  \item \textbf{ELM$_s$:} Extreme learning machine trained on source.
  \item \textbf{ELM$_t$:} Extreme learning machine trained on target.
  \item \textbf{GFK:} Geodesic Flow Kernel \cite{gong2012geodesic}.
  \item \textbf{MMDT:} Max-Margin Domain Transforms \cite{hoffman2013efficient,hoffman2014asymmetric}.
  \item \textbf{CDLS:} Cross-Domain Landmark Selection \cite{hubert2016learning}.
\end{itemize}


\begin{table}[htbp]
\centering
\caption{Summary of the Domains used in the experiments}
\label{tab1}
\begin{adjustbox}{max width=\columnwidth}
\begin{threeparttable}
\begin{tabular}{|c|c|c|c|c|c|c|}
\hline
Problem & Domains & Dataset & \# Samples & \# Features & \# Classes & Abbr.\tabularnewline
\hline
\hline
\multirow{4}{*}{Objects} & Amazon & Office & 958 & 800 & 10 & A\tabularnewline
\cline{2-7}
 & Webcam & Office & 295 & 800 & 10 & W\tabularnewline
\cline{2-7}
 & DSLR & Office & 157 & 800 & 10 & D\tabularnewline
\cline{2-7}
 & Caltech & Caltech-256 & 1,123 & 800 & 10 & C\tabularnewline
\hline
\multirow{5}{*}{Texts} & English & Multilingual & 18,758 & 11,547 & 6 & EN\tabularnewline
\cline{2-7}
 & French & Multilingual & 26,648 & 11,547 & 6 & FR\tabularnewline
\cline{2-7}
 & German & Multilingual & 29,953 & 11,547 & 6 & GR\tabularnewline
\cline{2-7}
 & Italian & Multilingual & 24,039 & 11,547 & 6 & IT\tabularnewline
\cline{2-7}
 & Spanish & Multilingual & 12,342 & 11,547 & 6 & SP\tabularnewline
\hline
\end{tabular}
\end{threeparttable}
\end{adjustbox}
\end{table}

\begin{table*}[htbp]
\centering
\caption{RECOGNITION ACCURACIES $(\%)$ ON THE Caltech-Office datasets with SURF feature}
\label{tab2}
\begin{adjustbox}{max width=\textwidth}
\begin{threeparttable}
\begin{tabular}{|c||c|c|c|c|c|c|c|c|c|c|c|c|c|}
\hline
Method & A$\rightarrow$C & A$\rightarrow$D & A$\rightarrow$W & C$\rightarrow$A & C$\rightarrow$D & C$\rightarrow$W & D$\rightarrow$A & D$\rightarrow$C & D$\rightarrow$W & W$\rightarrow$A & W$\rightarrow$C & W$\rightarrow$D & Mean\tabularnewline
\hline
\hline
SVM$_{S}$ & \textcolor{red}{38.6$\pm$0.4} & 33.4$\pm$1.3 & 34.8$\pm$0.8 & 38.5$\pm$0.6 & 33.9$\pm$1.0 & 30.2$\pm$1.0 & 36.4$\pm$0.5 & 32.8$\pm$0.3 & \textcolor{blue}{76.6$\pm$0.8} & 34.1$\pm$0.6 & 29.6$\pm$0.6 & 67.9$\pm$0.7 & 40.6$\pm$0.7\tabularnewline
\hline
SVM$_{T}$ & 34.2$\pm$0.6 & 55.5$\pm$0.8 & 63.1$\pm$0.8 & 47.0$\pm$1.1 & 55.3$\pm$1.1 & 59.4$\pm$1.4 & 46.5$\pm$1.0 & 33.4$\pm$0.6 & 60.3$\pm$1.2 & 48.5$\pm$0.9 & 31.1$\pm$0.8 & 53.5$\pm$1.0 & 49.0$\pm$0.9\tabularnewline
\hline
ELM$_{S}$ & \textcolor{blue}{36.8$\pm$0.4} & 31.2$\pm$1.2 & 31.0$\pm$1.1 & 38.1$\pm$0.7 & 35.2$\pm$1.0 & 30.3$\pm$1.3 & 36.5$\pm$0.6 & 30.7$\pm$0.5 & \textcolor{red}{78.2$\pm$0.5} & 32.7$\pm$0.7 & 29.1$\pm$0.5 & \textcolor{red}{72.8$\pm$0.9} & 40.2$\pm$0.8\tabularnewline
\hline
ELM$_{T}$ & 33.2$\pm$0.7 & 54.5$\pm$1.0 & \textcolor{blue}{65.5$\pm$1.1} & 48.8$\pm$0.9 & 56.6$\pm$0.8 & \textcolor{blue}{64.8$\pm$1.4} & 48.6$\pm$0.9 & 34.0$\pm$0.7 & 65.9$\pm$0.8 & \textcolor{blue}{49.9$\pm$1.0} & 31.4$\pm$0.9 & 57.6$\pm$0.8 & 50.9$\pm$0.9\tabularnewline
\hline
GFK & \textcolor{blue}{36.0$\pm$0.5} & 50.7$\pm$0.8 & 58.6$\pm$1.0 & 44.7$\pm$0.8 & \textcolor{red}{57.7$\pm$1.1} & 63.7$\pm$0.8 & 45.7$\pm$0.6 & 32.9$\pm$0.5 & \textcolor{blue}{76.5$\pm$0.5} & 44.1$\pm$0.4 & 31.1$\pm$0.6 & \textcolor{blue}{70.5$\pm$0.7} & 51.0$\pm$0.7 \tabularnewline
\hline
MMDT & \textcolor{blue}{36.4$\pm$0.8} & \textcolor{blue}{56.7$\pm$1.3} & 64.6$\pm$1.2 & \textcolor{blue}{49.4$\pm$0.8} & 56.5$\pm$0.9 & 63.8$\pm$1.1 & 46.9$\pm$1.0 & 34.1$\pm$0.8 & 74.1$\pm$0.8 & 47.7$\pm$0.9 & 32.2$\pm$0.8 & 64.0$\pm$0.7 & \textcolor{blue}{52.2$\pm$0.9 }\tabularnewline
\hline
CDLS & 28.7$\pm$1.0 & 54.4$\pm$1.3 & 60.5$\pm$1.1 & 41.0$\pm$1.0 & 53.2$\pm$1.1 & 61.6$\pm$0.9 & \textcolor{blue}{49.1$\pm$0.8} & \textcolor{blue}{35.7$\pm$0.6} & 75.1$\pm$0.8 & \textcolor{blue}{49.8$\pm$0.7} & \textcolor{red}{34.6$\pm$0.6} & 64.0$\pm$0.7 & 50.6$\pm$0.9 \tabularnewline
\hline
PTELM & \textcolor{blue}{36.0$\pm$0.7} & \textcolor{red}{57.0$\pm$0.8} & \textcolor{red}{67.0$\pm$0.8} & \textcolor{red}{51.2$\pm$0.9} & \textcolor{blue}{57.3$\pm$0.8} & \textcolor{red}{64.9$\pm$1.0} & \textcolor{red}{50.6$\pm$0.8} & \textcolor{red}{36.2$\pm$0.6} & 67.2$\pm$0.8 & \textcolor{red}{52.3$\pm$0.7} & \textcolor{blue}{33.5$\pm$0.9} & 59.2$\pm$0.8 & \textcolor{red}{52.7$\pm$0.8 }\tabularnewline
\hline
\end{tabular}
\footnotesize Red indicates the best result for each domain split. Blue indicates the group of results that are close to the best performing result. (A: \textit{Amazon}, C: \textit{Caltech}, D: \textit{DSLR} and W: \textit{Webcam})
\end{threeparttable}
\end{adjustbox}
\end{table*}

\begin{table*}[htbp]
\centering
\caption{RECOGNITION ACCURACIES $(\%)$ ON THE Multilingual Reuters Collection datasets with Spanish as target domain}
\label{tab3}
\begin{adjustbox}{max width=\textwidth}
\begin{threeparttable}
\begin{tabular}{|c||c|c|c|c|c|c|c|c||c|c|c|c|c|c|c|c|}
\hline
Source & \multicolumn{8}{c||}{\# labeled target domain data / category = 10} & \multicolumn{8}{c|}{\# labeled target domain data / category = 20}\tabularnewline
\cline{2-17}
Articles & SVM$_S$ & SVM$_T$ & ELM$_S$ & ELM$_T$ & GFK & MMDT & CDLS & PTELM & SVM$_S$ & SVM$_T$ & ELM$_S$ & ELM$_T$ & GFK & MMDT & CDLS & PTELM\tabularnewline
\hline
\hline
English & 28.8$\pm$1.3 & \multirow{4}{*}{68.5$\pm$1.0} & 39.9$\pm$1.6 & \multirow{4}{*}{67.0$\pm$1.0} & 64.2$\pm$0.7 & \textcolor{blue}{71.4$\pm$0.6} & 70.2$\pm$0.7 & \textcolor{red}{72.2$\pm$0.3} & 28.9$\pm$1.3 & \multirow{4}{*}{74.5$\pm$0.6} & 40.9$\pm$1.5 & \multirow{4}{*}{72.2$\pm$0.5} & 71.7$\pm$0.5 & 75.2$\pm$0.6 & \textcolor{blue}{76.5$\pm$0.5} & \textcolor{red}{77.2$\pm$0.3}\tabularnewline
\cline{1-2} \cline{4-4} \cline{6-10} \cline{12-12} \cline{14-17}
French & 53.0$\pm$0.9 &  & 56.6$\pm$1.0 &  & 66.9$\pm$0.6 & \textcolor{blue}{72.8$\pm$0.4} & 70.5$\pm$0.8 & \textcolor{red}{73.1$\pm$0.5} & 52.6$\pm$0.9 &  & 58.3$\pm$0.8 &  & 72.3$\pm$0.6 & 74.7$\pm$0.4 & \textcolor{blue}{75.6$\pm$0.6} & \textcolor{red}{76.7$\pm$0.3}\tabularnewline
\cline{1-2} \cline{4-4} \cline{6-10} \cline{12-12} \cline{14-17}
German & 39.1$\pm$1.2 &  & 48.2$\pm$0.8 &  & 65.2$\pm$0.7 & \textcolor{blue}{72.1$\pm$0.6} & 70.8$\pm$0.8 & \textcolor{red}{73.8$\pm$0.4} & 39.0$\pm$1.2 &  & 46.0$\pm$1.3 &  & 70.8$\pm$0.5 & \textcolor{blue}{75.7$\pm$0.5} & \textcolor{blue}{75.9$\pm$0.5} & \textcolor{red}{77.2$\pm$0.3}\tabularnewline
\cline{1-2} \cline{4-4} \cline{6-10} \cline{12-12} \cline{14-17}
Italian & 63.5$\pm$0.6 &  & 56.9$\pm$1.0 &  & 65.7$\pm$0.7 & \textcolor{blue}{72.5$\pm$0.6} & 71.0$\pm$0.9 & \textcolor{red}{73.3$\pm$0.5} & 63.2$\pm$0.6 &  & 57.9$\pm$0.7 &  & 71.6$\pm$0.6 & \textcolor{red}{76.2$\pm$0.5} & \textcolor{blue}{75.9$\pm$0.5} & \textcolor{red}{76.2$\pm$0.4}\tabularnewline
\hline
Mean & 46.1$\pm$1.0 & 68.5$\pm$1.0 & 50.4$\pm$1.1 & 67.0$\pm$1.0 & 65.5$\pm$0.7 & \textcolor{blue}{72.2$\pm$0.6} & 70.6$\pm$0.8 & \textcolor{red}{73.1$\pm$0.4} & 45.9$\pm$1.0 & 74.5$\pm$0.6 & 50.8$\pm$1.1 & 72.2$\pm$0.5 & 71.6$\pm$0.6 & 75.5$\pm$0.5 & \textcolor{blue}{76.0$\pm$0.5} & \textcolor{red}{76.8$\pm$0.3}\tabularnewline
\hline
\end{tabular}
\footnotesize Red indicates the best result for each domain split. Blue indicates the group of results that are close to the best performing result.
\end{threeparttable}
\end{adjustbox}
\end{table*}

\subsection{Cross-Domain Object Recognition}
For our first experiment, we use the \emph{Caltech-Office} domain adaptation benchmark dataset to evaluate our method on the real world computer vision adaptation tasks.
\subsubsection{Experiment Setup} Following the setup of \cite{gong2012geodesic,saenko2010adapting,hoffman2013efficient}, the number of selected labeled source samples per class for \textit{amazon}, \textit{webcam}, \textit{dslr} and \textit{caltech} is 20, 8, 8, and 8, respectively. Instead, when they serve as target domain, 3 labeled target samples are used. We use the same 20 random train/test splits download from the website\footnote{\url{https://people.eecs.berkeley.edu/~jhoffman/domainadapt/}} provided by the authors \cite{hoffman2013efficient} for fair comparison and report averaged results across them.

For our method, we fix $\lambda_1=1$, $\lambda_2=30$ and $\lambda_3=10$. The number of hidden nodes of the ELM networks is set as 500 in all experiments. For other baseline methods, we use the recommended parameters.

\subsubsection{Results}
We report the mean and standard deviation of classification accuracies for all methods on the \textit{Office-Caltech} dataset in Table \ref{tab2}. Each result in the same column is based on the same 20 random trials. As can be seen, our proposed method outperforms all other methods in 7 out of the 12 individual domain shifts and achieves the highest average accuracy 52.7\% over the all 12 domain shift experiments. It is worth noticing that our PTELM typically outperforms the other competing methods when \textit{amazon} serve as source or target domain. We believe the reason is that the domain discrepancy between \textit{amazon}$\rightarrow$\textit{webcam} and \textit{amazon}$\rightarrow$\textit{dslr} are much more significant than other domain shifts, as the larger performance discrepancy between the ELM$_s$ and ELM$_t$ in these domain shifts. Therefore, it is obvious that our approach is more effective to deal with large domain shifts.

\begin{figure*}[htbp]
\begin{center}
  \maxsizebox{0.89\textwidth}{!}{\includegraphics{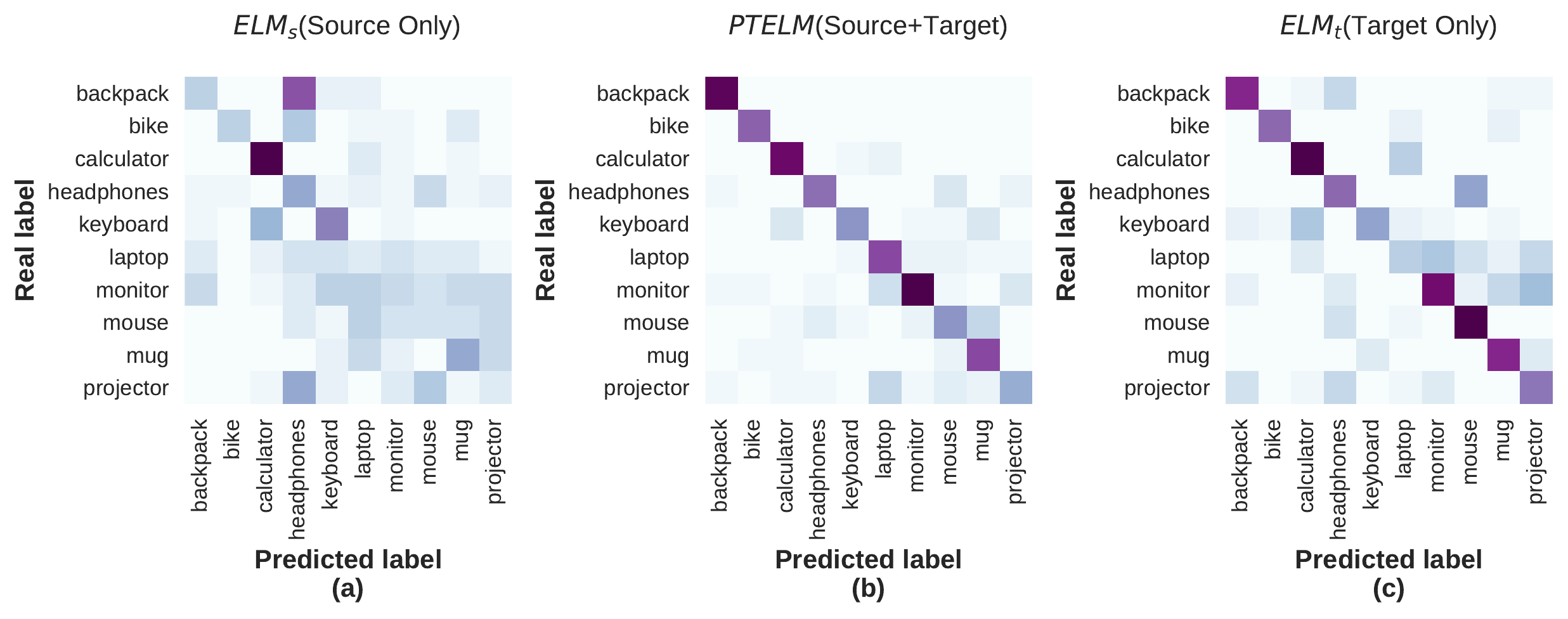}}
\end{center}
\caption{Confusion matrices of the \textit{amazon}$\rightarrow$\textit{webcam} domain shift experiment. Left: ELM model trained with source domain only. Middle: Our proposed PTELM method trained with source and target domain together. Right: ELM model trained with target domain only.}
\label{fig1}
\end{figure*}

We also visualize the effectiveness of the proposed PTELM via the confusion matrix. Figure \ref{fig1} plots the confusion matrices of ELM$_s$, PTELM and ELM$_t$ on \textit{amazon}$\rightarrow$\textit{webcam} domain shift experiment. By inspecting the confusion matrix of ELM$_s$, which trained with 20 labeled source samples per class, we find that the source only model is heavily confused about several classes. It also reveals the large domain shift between \textit{amazon} and \textit{webcam} and gives explanation for the performance discrepancy between ELM$_s$ and ELM$_t$. On the other hand, the confusion matrix of ELM$_t$, which trained with 3 labeled target samples per class, is also somewhat confused. In contrast, as can be seen in Figure \ref{fig1}(b), the off-diagonal elements in confusion matrix are close to zero, which demonstrates that our PTELM method can effectively utilize source and target damain samples together to train a high-quality classifier.

\subsection{Cross-Domain Text Categorization}
For the second experiment, we utilize the Multilingual Reuters Collection dataset to evaluate our method in the context of text categorization.
\subsubsection{Experiment Setup} In this dataset, documents written in different languages can be viewed as different domains. We take \textit{Spanish} as target domain, and other four languages (\textit{English}, \textit{French}, \textit{German} and \textit{Italian}) as individual source domain. Therefore, there are four combinations in total. For each category, we randomly sample 100 labeled training documents from source domain and $m$ labeled training documents from target domain, where $m =$ 5, 10, 15 and 20, respectively. And the remaining documents in the target domain are used as the test set\footnote{The splits we used can be downloaded from \url{https://github.com/BoyuanJiang/PTELM/tree/master/DataSplits}}. Note that the dimensions of the original TF-IDF features are up to 11,547, in order to fairly compare our method with other competing methods, we perform principal components analysis\footnote{The PCA uses randomized singular value decomposition algorithm as SVD solver for efficiency.} for dimension reduction and the dimensions after PCA are 40.

In this experiment, we also fix $\lambda_1=1$, $\lambda_2=30$ and $\lambda_3=10$. The number of hidden nodes is set as $L=600$ instead.

\subsubsection{Results}

\begin{figure*}[htbp]
\begin{center}
  \maxsizebox{0.89\textwidth}{!}{\includegraphics{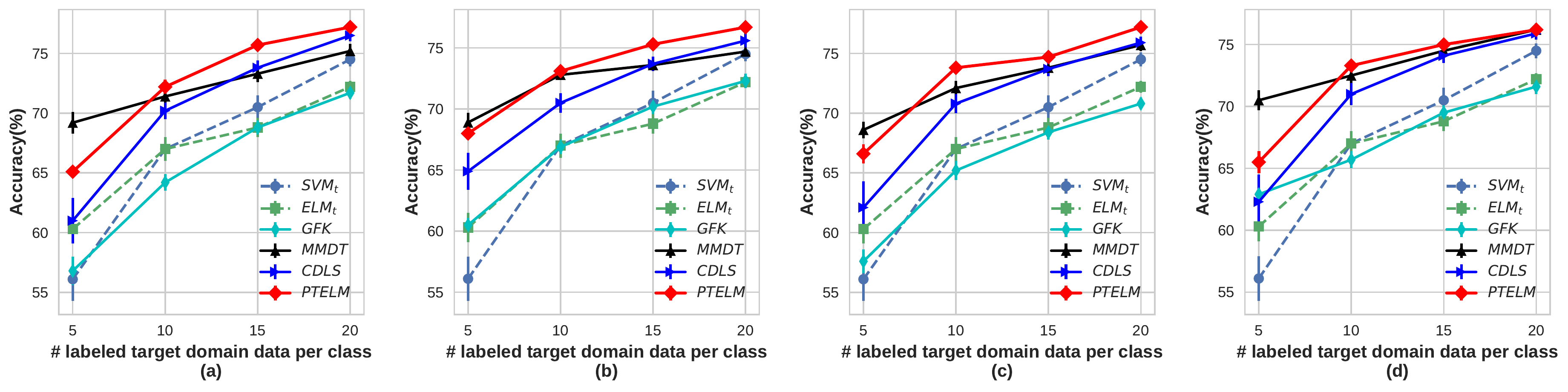}}
\end{center}
\caption{Classification accuracies of all methods with varied labeled target data per class (i.e. $m$ = 5, 10, 15 and 20) on the Multilingual Reuters Collection dataset. Note that \textit{Spanish} is considered as target domain, while the source domains are selected from (a) \textit{English}, (b) \textit{French}, (c) \textit{German} and (d) \textit{Italian}, respectively}
\label{fig2}
\end{figure*}

\begin{figure}[htbp]
\begin{center}
\maxsizebox{0.89\columnwidth}{!}{\includegraphics{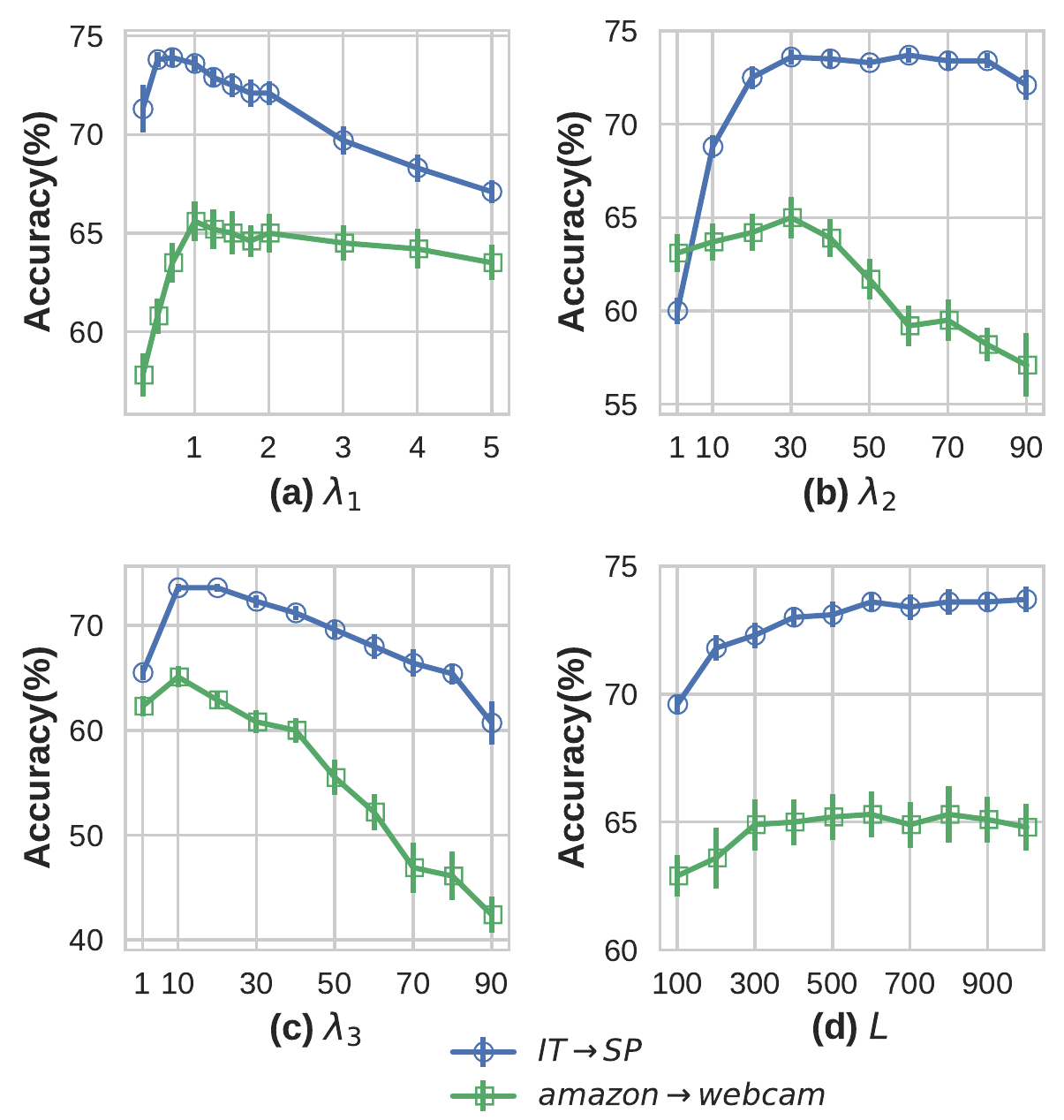}}
\end{center}
\caption{Parameter sensitivity study for the PTELM algoritnm on \textit{IT}$\rightarrow$\textit{SP} and \textit{amazon}$\rightarrow$\textit{webcam} domain shifts.}
\label{sensitivity}
\end{figure}
We report means and standard deviations of all methods on the Multilingual Reuters Collection dataset when $m = 10$ and $20$ in Table \ref{tab3}. It is obvious that our proposed PTELM method beats other competing methods under both settings. It is interesting to note that the GKF algorithm works worse than the ELM$_t$ and SVM$_t$. A possible explanation is that the GFK is put forward for unsupervised domain adaptation, therefore, does not utilize the given target label for training.

We also plot means and standard deviations of all methods over different number of labeled target samples (5, 10, 15 and 20 respectively) in Figure \ref{fig2} except SVM$_s$ and ELM$_s$, as these two methods perform much worse than the other methods. From the figure, it can be seen that the performance of all the methods is improved with the increase of the number of labeled target samples and our method performs best in most cases. It is worth noting that MMDT performs a little better than our method and much better than other methods when $m = 5$, which demonstrates that MMDT is more suitable when few labeled target samples are available. Besides, another key insight from the figure is that our method is more stable than the competing methods with lower standard deviations.

\subsection{Parameter Sensitivity}
In this section, we investigate the sensitivity of four parameters involved in our method, which are three trade-off parameters $\lambda_1,\lambda_2,\lambda_3$ and the number of hidden nodes $L$, respectively. Due to space limitation, we only choose \textit{amazon}$\rightarrow$\textit{webcam} from the \textit{Office-Caltech} dataset and \textit{IT}$\rightarrow$\textit{SP} from the Multilingual Reuters Collection dataset to evaluate accuracy. Each time, only one parameter is allowed to change with the other parameters fixed. The results are shown in Figure \ref{sensitivity} and we give a brief analysis here. For $\lambda_1$, it is the trade-off parameter to balance the contribution of the source and target domain. When $\lambda_1$ is smaller than 1, the model learns more from the source domain. On the contrary, when $\lambda_1$ is larger than 1, the target domain counts more. Therefore, a reasonable value of $\lambda_1$ is close to 1, as can be seen in Figure \ref{sensitivity} (a). $\lambda_2$ and $\lambda_3$ are two penalty terms to prevent the model from overfitting the source and target domain data. As can be seen in Figure \ref{sensitivity} (b) and (c), the reasonable choices could be $\lambda_2\in[20,40]$ and $\lambda_3\in[10,30]$. For the number of hidden nodes $L$, it is highly related to feature dimensions and a reasonable value is about 500 in our experiments.

\section{Conclusion and Future Work}
In this paper, we presented a novel approach for parameter transfer under the ELM framework, which explicitly bridges the source domain parameters and the target domain parameters by a projection matrix. In order to select informative source domain features for knowledge transfer, the $\ell_{21}\text{-norm}$ was applied to the source parameters. Additionally, an effective alternate optimization method was introduced to jointly learn the projection matrix and the model parameters. Experiments on several challenging datasets showed that the proposed PTELM significantly outperforms the non-transfer ELM and SVM by a large margin, besides, achieves better performance than the other representative methods.
\\\indent In the future, we plan to extend our proposal in the following two aspects. (1) Extending the PTELM to multiple source domain adaptation method. (2) Reformulating the model by transforming the source and target parameters into a shared parameter space by two different projection matrices.

\bibliographystyle{IEEEtran}
\bibliography{IEEEabrv,paper}

\begin{thebibliography}{10}
\providecommand{\url}[1]{#1}
\csname url@samestyle\endcsname
\providecommand{\newblock}{\relax}
\providecommand{\bibinfo}[2]{#2}
\providecommand{\BIBentrySTDinterwordspacing}{\spaceskip=0pt\relax}
\providecommand{\BIBentryALTinterwordstretchfactor}{4}
\providecommand{\BIBentryALTinterwordspacing}{\spaceskip=\fontdimen2\font plus
\BIBentryALTinterwordstretchfactor\fontdimen3\font minus
  \fontdimen4\font\relax}
\providecommand{\BIBforeignlanguage}[2]{{%
\expandafter\ifx\csname l@#1\endcsname\relax
\typeout{** WARNING: IEEEtran.bst: No hyphenation pattern has been}%
\typeout{** loaded for the language `#1'. Using the pattern for}%
\typeout{** the default language instead.}%
\else
\language=\csname l@#1\endcsname
\fi
#2}}
\providecommand{\BIBdecl}{\relax}
\BIBdecl

\bibitem{pan2010survey}
S.~J. Pan and Q.~Yang, ``A survey on transfer learning,'' \emph{IEEE
  Transactions on knowledge and data engineering}, vol.~22, no.~10, pp.
  1345--1359, 2010.

\bibitem{long2014transfer}
M.~Long, J.~Wang, G.~Ding, J.~Sun, and P.~S. Yu, ``Transfer joint matching for
  unsupervised domain adaptation,'' pp. 1410--1417, 2014.

\bibitem{long2014adaptation}
M.~Long, J.~Wang, G.~Ding, S.~J. Pan, and S.~Y. Philip, ``Adaptation
  regularization: A general framework for transfer learning,'' \emph{IEEE
  Transactions on Knowledge and Data Engineering}, vol.~26, no.~5, pp.
  1076--1089, 2014.

\bibitem{zhang2016robust}
L.~Zhang and D.~Zhang, ``Robust visual knowledge transfer via extreme learning
  machine-based domain adaptation,'' \emph{IEEE Transactions on Image
  Processing}, vol.~25, no.~10, pp. 4959--4973, 2016.

\bibitem{liu2017common}
Y.~Liu, L.~Zhang, P.~Deng, and Z.~He, ``Common subspace learning via
  cross-domain extreme learning machine,'' \emph{Cognitive Computation}, pp.
  1--9, 2017.

\bibitem{hoffman2014asymmetric}
J.~Hoffman, E.~Rodner, J.~Donahue, B.~Kulis, and K.~Saenko, ``Asymmetric and
  category invariant feature transformations for domain adaptation,''
  \emph{International journal of computer vision}, vol. 109, no. 1-2, pp.
  28--41, 2014.

\bibitem{saenko2010adapting}
K.~Saenko, B.~Kulis, M.~Fritz, and T.~Darrell, ``Adapting visual category
  models to new domains,'' \emph{Computer Vision--ECCV 2010}, pp. 213--226,
  2010.

\bibitem{gong2012geodesic}
B.~Gong, Y.~Shi, F.~Sha, and K.~Grauman, ``Geodesic flow kernel for
  unsupervised domain adaptation,'' pp. 2066--2073, 2012.

\bibitem{hoffman2013efficient}
J.~Hoffman, E.~Rodner, J.~Donahue, T.~Darrell, and K.~Saenko, ``Efficient
  learning of domain-invariant image representations,'' \emph{international
  conference on learning representations}, 2013.

\bibitem{kulis2011you}
B.~Kulis, K.~Saenko, and T.~Darrell, ``What you saw is not what you get: Domain
  adaptation using asymmetric kernel transforms,'' pp. 1785--1792, 2011.

\bibitem{sun2016return}
B.~Sun, J.~Feng, and K.~Saenko, ``Return of frustratingly easy domain
  adaptation,'' \emph{national conference on artificial intelligence}, pp.
  2058--2065, 2016.

\bibitem{chen2018joint}
C.~Chen, Z.~Chen, B.~Jiang, and X.~Jin, ``Joint domain alignment and
  discriminative feature learning for unsupervised deep domain adaptation,''
  \emph{arXiv preprint arXiv:1808.09347}, 2018.

\bibitem{yang2007adapting}
J.~Yang, R.~Yan, and A.~G. Hauptmann, ``Adapting svm classifiers to data with
  shifted distributions,'' pp. 69--76, 2007.

\bibitem{aytar2011tabula}
Y.~Aytar and A.~Zisserman, ``Tabula rasa: Model transfer for object category
  detection,'' pp. 2252--2259, 2011.

\bibitem{tommasi2014learning}
T.~Tommasi, F.~Orabona, and B.~Caputo, ``Learning categories from few examples
  with multi model knowledge transfer,'' \emph{IEEE transactions on pattern
  analysis and machine intelligence}, vol.~36, no.~5, pp. 928--941, 2014.

\bibitem{li2016extreme}
X.~Li, W.~Mao, and W.~Jiang, ``Extreme learning machine based transfer learning
  for data classification,'' \emph{Neurocomputing}, vol. 174, pp. 203--210,
  2016.

\bibitem{huang2006extreme}
G.-B. Huang, Q.-Y. Zhu, and C.-K. Siew, ``Extreme learning machine: theory and
  applications,'' \emph{Neurocomputing}, vol.~70, no.~1, pp. 489--501, 2006.

\bibitem{huang2012extreme}
G.-B. Huang, H.~Zhou, X.~Ding, and R.~Zhang, ``Extreme learning machine for
  regression and multiclass classification,'' \emph{IEEE Transactions on
  Systems, Man, and Cybernetics, Part B (Cybernetics)}, vol.~42, no.~2, pp.
  513--529, 2012.

\bibitem{huang2014semi}
G.~Huang, S.~Song, J.~N. Gupta, and C.~Wu, ``Semi-supervised and unsupervised
  extreme learning machines,'' \emph{IEEE transactions on cybernetics},
  vol.~44, no.~12, pp. 2405--2417, 2014.

\bibitem{chen2018optimizing}
C.~Chen, X.~Jin, B.~Jiang, and L.~Li, ``Optimizing extreme learning machine via
  generalized hebbian learning and intrinsic plasticity learning,''
  \emph{Neural Processing Letters}, pp. 1--17, 2018.

\bibitem{zong2013weighted}
W.~Zong, G.-B. Huang, and Y.~Chen, ``Weighted extreme learning machine for
  imbalance learning,'' \emph{Neurocomputing}, vol. 101, pp. 229--242, 2013.

\bibitem{liang2006fast}
N.-Y. Liang, G.-B. Huang, P.~Saratchandran, and N.~Sundararajan, ``A fast and
  accurate online sequential learning algorithm for feedforward networks,''
  \emph{IEEE Transactions on neural networks}, vol.~17, no.~6, pp. 1411--1423,
  2006.

\bibitem{zhou2015stacked}
H.~Zhou, G.-B. Huang, Z.~Lin, H.~Wang, and Y.~C. Soh, ``Stacked extreme
  learning machines,'' \emph{IEEE transactions on cybernetics}, vol.~45, no.~9,
  pp. 2013--2025, 2015.

\bibitem{huang2015local}
G.-B. Huang, Z.~Bai, L.~L.~C. Kasun, and C.~M. Vong, ``Local receptive fields
  based extreme learning machine,'' \emph{IEEE Computational Intelligence
  Magazine}, vol.~10, no.~2, pp. 18--29, 2015.

\bibitem{zhang2015domain}
L.~Zhang and D.~Zhang, ``Domain adaptation extreme learning machines for drift
  compensation in e-nose systems,'' \emph{IEEE Transactions on instrumentation
  and measurement}, vol.~64, no.~7, pp. 1790--1801, 2015.

\bibitem{uzair2017blind}
M.~Uzair and A.~Mian, ``Blind domain adaptation with augmented extreme learning
  machine features,'' \emph{IEEE transactions on cybernetics}, vol.~47, no.~3,
  pp. 651--660, 2017.

\bibitem{salaken2017extreme}
S.~M. Salaken, A.~Khosravi, T.~Nguyen, and S.~Nahavandi, ``Extreme learning
  machine based transfer learning algorithms: A survey,''
  \emph{Neurocomputing}, vol. 267, pp. 516--524, 2017.

\bibitem{ding2006r}
C.~H.~Q. Ding, D.~Zhou, X.~He, and H.~Zha, ``R1-pca: rotational invariant
  $\ell_1\text{-norm}$ principal component analysis for robust subspace
  factorization,'' pp. 281--288, 2006.

\bibitem{gu2011joint}
Q.~Gu, Z.~Li, and J.~Han, ``Joint feature selection and subspace learning,''
  pp. 1294--1299, 2011.

\bibitem{nie2010efficient}
F.~Nie, H.~Huang, X.~Cai, and C.~H.~Q. Ding, ``Efficient and robust feature
  selection via joint $\ell_{2,1}\text{-Norms}$ minimization,'' pp. 1813--1821,
  2010.

\bibitem{hubert2016learning}
Y.~H. Tsai, Y.~Yeh, and Y.~F. Wang, ``Learning cross-domain landmarks for
  heterogeneous domain adaptation,'' pp. 5081--5090, 2016.

\bibitem{griffin2007caltech}
G.~Griffin, A.~Holub, and P.~Perona, ``Caltech-256 object category dataset,''
  2007.

\bibitem{bay2006surf}
H.~Bay, T.~Tuytelaars, and L.~Van~Gool, ``Surf: Speeded up robust features,''
  \emph{Computer vision--ECCV 2006}, pp. 404--417, 2006.

\bibitem{AUG09}
M.-R. Amini, N.~Usunier, and C.~Goutte, ``Learning from multiple partially
  observed views - an application to multilingual text categorization,'' in
  \emph{NIPS 22}, 2009.

\bibitem{USLJ07}
N.~Ueffing, M.~Simard, S.~Larkin, and H.~Johnson, ``Nrc’s portage system for
  wmt 2007,'' \emph{ACL 2007}, pp. 185--188, 2007.

\end{thebibliography}
\end{document}